\documentclass[twoside]{article}
\usepackage[accepted]{aistats2021}
\usepackage{amsmath}
\usepackage{dsfont}
\usepackage{booktabs}
\usepackage{graphicx}
\newcommand{\indep}{\raisebox{0.05em}{\rotatebox[origin=c]{90}{$\models$}}}
\usepackage{amsthm}
\usepackage{amssymb}
\usepackage{url}
\usepackage{enumitem}

\usepackage{algorithmic}
\newcommand{\algcaption}[1]{%
  \par\noindent
  \textbf{Algorithm:}
  #1\par
}

\newtheorem{theorem}{Theorem}

\newtheorem{assumption}{Assumption}

\newtheorem{corollary}{Corollary}

\newtheorem*{remark}{Remark}

\usepackage{xcolor}
\usepackage[round]{natbib}

\begin{document}

%

%

\twocolumn[

\aistatstitle{Nonparametric Estimation of Heterogeneous Treatment Effects: From Theory to Learning Algorithms}

\aistatsauthor{Alicia Curth \And Mihaela van der Schaar}

\aistatsaddress{University of Cambridge \\ \url{amc253@cam.ac.uk} \And University of Cambridge,  UCLA \& The Alan Turing Institute  \\ \url{mv472@cam.ac.uk}} ]

\begin{abstract}
The need to evaluate treatment effectiveness is ubiquitous in most of empirical science, and interest in  flexibly investigating effect heterogeneity is growing rapidly. To do so, a multitude of model-agnostic, nonparametric meta-learners have been proposed in recent years. Such learners decompose the treatment effect estimation problem into separate sub-problems, each solvable using standard supervised learning methods. Choosing between different meta-learners in a data-driven manner is difficult, as it requires access to counterfactual information. Therefore, with the ultimate goal of building better understanding of the conditions under which some learners can be expected to perform better than others \textit{a priori}, we theoretically analyze four broad meta-learning strategies which rely on plug-in estimation and pseudo-outcome regression. We highlight how this theoretical reasoning can be used to guide principled algorithm design and translate our analyses into practice by considering a variety of neural network architectures as base-learners for the discussed meta-learning strategies. In a simulation study, we showcase the relative strengths of the learners under different data-generating processes. 
\end{abstract}


\section{INTRODUCTION}
Many empirical scientists ultimately aim to assess the causal effects of interventions, policies and treatments by analyzing experimental or observational data using tools from applied statistics. Due to the impressive performance of machine learning (ML) methods on prediction tasks, recent years have seen exciting developments incorporating ML into the estimation of average treatment effects \citep{van2011targeted, chernozhukov2017double}. While average treatment effects (ATEs) have been the main estimand of interest thus far, data-adaptive, ML-based estimators have even more potential to shape our ability to flexibly investigate heterogeneity of effects across populations. As interest moves towards personalized policy- and treatment design in fields such as econometrics and medicine, the need to accurately estimate the full conditional average treatment effect (CATE) function becomes ubiquitous.

The causal inference communities across disciplines have produced a rapidly growing number of algorithms for CATE estimation in recent years (see e.g. \cite{bica2020real} for an overview). In practice, this leads to the need to select the best model -- which is notoriously difficult in treatment effect studies because the ground truth is unobserved.  While recent literature has presented promising solutions using data-driven strategies \citep{rolling2014model, alaa2019validating}, we  believe that it is equally important to reduce the complexity of the selection task a priori by building greater systematic understanding of the strengths and weaknesses of different algorithms from a theoretical viewpoint.  


Here, we put our focus on comparing different so-called meta-learners for binary treatment effect estimation, which are model-agnostic algorithms that decompose the task of estimating CATE into multiple sub-problems, each solvable using \textit{any} supervised learning/ regression method \citep{kunzel2019metalearners}. In the theoretical part of this paper (Sections 3 and 4), we consider estimation within a generic nonparametric regression framework, i.e. we assume no known parametric structure, and derive theoretical arguments why one learner may outperform others. In the more practical part (Sections 5 and 6), we compare the empirical performance of the different learners using \textit{the same} underlying machine learning method, and consider a variety of neural network (NN) architectures for CATE estimation. Throughout, instead of arguing that one learning algorithm is superior to all others, we aim to highlight how expert knowledge on the underlying data-generating process (DGP) can narrow the choice of algorithms a priori and guide model design.

\begin{figure}[t]
\centering
\includegraphics[width=\columnwidth]{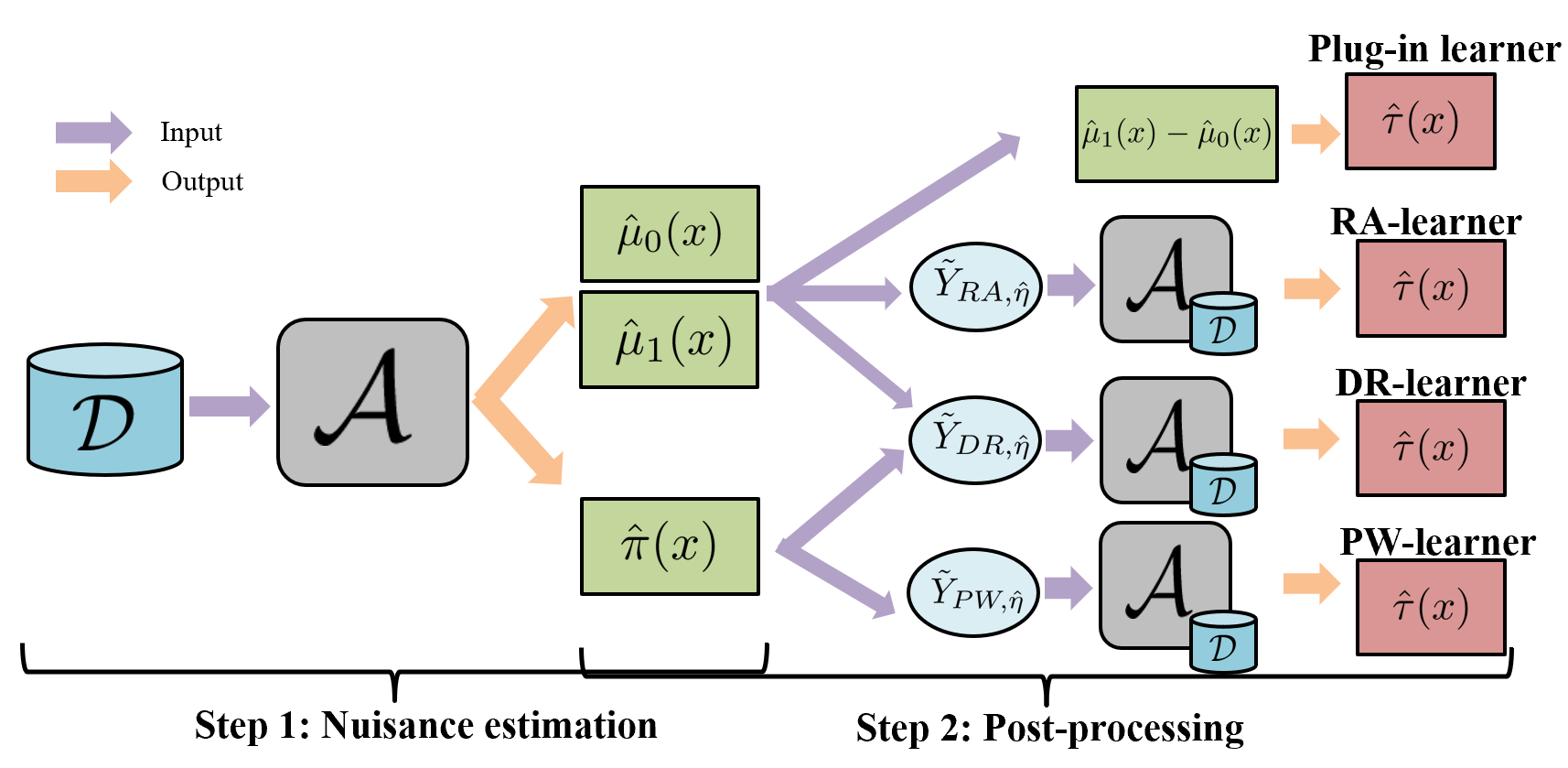}
\caption{High-level overview of the four meta-learning strategies considered in this paper. $\mathcal{A}$ refers to a generic regression algorithm and $\mathcal{D}$ refers to input data, both of which are used to estimate the propensity score $\hat{\pi}(x)$, the potential outcomes $\hat{\mu}_w(x)$ and CATE $\hat{\tau}(x)$, and to compute the pseudo-outcomes $\tilde{Y}_{l, \hat{\eta}}$.}\label{overviewfig}
\end{figure}

\textbf{Contributions} Our contributions are three-fold: First, we provide theoretical insights into nonparametric CATE estimation using meta-learners. We propose a new classification of meta-learners inspired by the ATE estimator taxonomy, categorizing algorithms into four broad classes: one-step plug-in learners and three types of two-step learners, which use unbiased pseudo-outcomes based on regression adjustment (RA), propensity weighting (PW) or both (DR), as illustrated in Figure \ref{overviewfig}. We present an analysis of the theoretical properties of the learners and discuss resulting theoretical criteria for choosing between them. While both plug-in and DR-learner have been previously analyzed, our analysis and discussion of RA- and PW-learner are -- to the best of our knowledge -- new.\\
Second, we compare four existing model architectures for CATE estimation using NNs and propose a new architecture which generalizes existing approaches. These architectures allow for different degrees of information sharing between nuisance parameter estimators, and we highlight the (dis-)advantages of different architectures. We also provide a suite of sklearn-style implementations for all architectures and meta-learners we consider\footnote{The code is available at \url{https://github.com/AliciaCurth/CATENets} and at \url{https://github.com/vanderschaarlab/mlforhealthlabpub/tree/main/alg/CATENets}}.\\
Third, we illustrate our theoretical arguments in simulation experiments, demonstrating how differences in DGPs and sample size influence the relative performance of different learners empirically. By considering how to best combine model architectures and meta-learner strategies, we also attempt to bridge the gap between the relatively disjoint literatures on meta-learners and end-to-end CATE estimation using NNs.

\subsection{Related Work}
We restrict our attention to so-called `meta-learners' for CATE -- model-agnostic algorithms that can be implemented using \textit{any arbitrary} ML method. \cite{kunzel2019metalearners} appear to be the first paper explicitly discussing meta-learning strategies for CATE estimation, of which they consider and named three in detail: The S-learner (\textit{single} learner), in which the treatment indicator is simply included as an additional feature in otherwise standard regression, the T-learner (\textit{two} learners) which fits separate regression functions for each treatment group and then takes differences, and the X-learner, a two-step regression estimator that uses each observation twice (see discussion in Section \ref{theory}). \cite{nie2017quasi} propose a two-step algorithm that estimates CATE using orthogonalization with respect to the nuisance functions, which they dub R-learner as it is based on \cite{robinson1988root}. Finally, \cite{kennedy2020optimal} proposes the DR-learner (\textit{doubly robust} learner), a two-step algorithm that uses the expression for the doubly robust augmented inverse propensity weighted (AIPW) estimator \citep{robins1995semiparametric} as a pseudo-outcome in a two-step regression set-up. With the exception of the DR-learner, the current naming strategy of meta-learners is surprisingly disjoint from the naming of ATE estimators, resulting in names that do not necessarily reflect the statistical concepts the learners are based on. To build better intuition and to facilitate principled theoretical analyses, we re-categorize meta-learning strategies in Section \ref{theory}.

Our theoretical considerations build on the statistical analyses of the fundamental limits of CATE estimation presented for Bayesian nonparametrics in \cite{alaa2018bayesian, alaa2018limits}, as well as the analyses of frequentist estimation using S-, T- and X-learner in \cite{kunzel2019metalearners} and the DR-learner in \cite{kennedy2020optimal}. All prior work considers only a subset of (variations of) the meta-learners we consider here, and can hence not lead to comprehensive practical advice on choosing between all strategies. 

While our theoretical analyses of meta-learners apply to generic estimators, in the practical part of this paper we focus on instantiations using standard feed-forward neural networks (NNs) and NN-based representation learning. This choice is motivated by interesting recent implementations \citep{johansson2016learning, shalit2017estimating, hassanpour2019learning, shi2019adapting} of hybrids of \cite{kunzel2019metalearners}'s S- and T-learners,  which build on ideas from  representation learning \citep{bengio2013representation} and multi-task learning \citep{caruana1997multitask}, sharing some information between regression tasks. We discuss these in detail in Section \ref{implementation}.  As the meta-learners can be used with arbitrary regression estimators, a wide range of other ML methods have been used in practice, and popular choices include Bayesian Additive Regression Trees \citep{Hill2011} and random forests \citep{kunzel2019metalearners}. Note that, in contrast to the model-agnostic nature of the meta-learners, there also exist a variety of CATE estimators that rely on a \textit{specific ML method} -- e.g. generative adversarial networks \citep{yoon2018ganite}, deep kernel learning \citep{zhang2020learning} or generalized random forests \citep{Athey2019} -- which therefore fall outside the scope of this paper.

\section{PROBLEM DEFINITION}
Assume we observe a sample $\mathcal{D}=\{(Y_i, X_i, W_i)\}^n_{i=1}$, with $(Y_i, X_i, W_i) \stackrel{i.i.d.}{\sim} \mathbb{P}$. Here, $Y \in \mathcal{Y}$ is a continuous or binary outcome of interest, $X_i \in \mathcal{X} \subset \mathbb{R}^d$ a $d$-dimensional covariate vector of possible confounders and $W_i\in \{0, 1\}$ is a binary treatment, which is assigned according to propensity score $\pi(x) = \mathbb{P}(W=1| X=x)$, with marginal treatment assignment probability $p_\pi=\mathbb{P}(W=1)$. Using the Neyman-Rubin potential outcomes framework \citep{rubin2005causal}, our main interest lies in the individualized treatment effect: the difference between the potential outcomes $Y_i(0)$ if individual $i$ does not receive treatment ($W_i=0$) and $Y_i(1)$ if treatment is administered ($W_i=1$).  However, by the \textit{fundamental problem of causal inference}, only one of the potential outcomes is observed, since $Y_i = W_i Y_i(1) + (1-W_i)Y_i(0)$. Therefore, in line with the majority of existing literature, we focus on estimating the conditional average treatment effect (CATE), 
\begin{equation}
\tau(x) = \mathbb{E}_{\mathbb{P}}[Y(1) - Y(0) |X =x]
\end{equation}
the expected treatment effect for an individual with covariate values $X=x$. 

It is well known that the identification of causal effects from observational data hinges on the imposition of untestable assumptions. Here, we consider estimation under the standard assumptions:

\begin{assumption}\label{causass}[Consistency, unconfoundedness and overlap]
Consistency: If individual i is assigned treatment $w_i$, we observe the associated potential outcome $Y_i=Y_i(w_i)$. 
Unconfoundedness: there are no unobserved confounders, such that $Y(0), Y(1) \indep W | X$. Overlap: treatment assignment is non-deterministic, i.e. $0 < \pi(x) < 1 \text{, }\forall x\in \mathcal{X}$. 
\end{assumption}
Under assumption \ref{causass}, CATE can be written as $\tau(x) = \mu_1(x) - \mu_0(x)$ for $\mu_w(x) = \mathbb{E}_{\mathbb{P}}[Y|W=w, X=x]$, and can be estimated from observational data using the meta-learners we discuss. We assume no known parametric form for $\tau(x)$, any of the  nuisance parameters $\eta=(\mu_0(x), \mu_1(x), \pi(x))$ or error distributions in $\mathbb{P}$, leaving us with a \textit{nonparametric} estimation problem. Throughout the theoretical analysis we consider generic nonparametric regression estimators, and use NNs in the experiments, yet all strategies could be used directly with other methods, e.g. random forests.

\section{CATEGORIZING CATE META-LEARNERS}\label{theory}
To improve our ability to analyze the meta-learner`s theoretical performance in a structured manner in the following section, we propose a high-level classification of CATE meta-learners that follows the well-known ATE taxonomy of estimators (see e.g. \cite{imbens2004nonparametric}). We prefer this naming strategy because it builds on existing intuition and classifies learners by the characteristics that reflect their statistical properties. Therefore, we suggest to divide meta-learners into \textit{one-step plug-in learners} -- learners that output two regression functions which can then be differenced -- and \textit{two-step learners}, based on a \textit{regression adjustment (RA)}, \textit{propensity weighting (PW)} or \textit{doubly robust (DR)} strategy, outputting a CATE function directly. These strategies are illustrated in Figure \ref{overviewfig}. 

We define one-step plug-in learners as those who obtain regression functions $\hat{\mu}_w(x)$ from the observed data, and estimate CATE directly as $\hat{\tau}(x) = \hat{\mu}_1(x) - \hat{\mu}_0(x)$. This is the strategy underlying \cite{kunzel2019metalearners}'s S- and T-learners. We do consider it important to keep a distinction between S- and T-learner as special cases of plug-in learners, because they differ in an important dimension: namely the amount of information shared between the potential outcome estimators. We use these terms to summarize somewhat broader estimator classes than \cite{kunzel2019metalearners}: We consider as T-learners those that separate estimation of the nuisance functions into disjoint sub-tasks (of which are \textit{t}wo or \textit{t}hree, depending on estimation of $\pi(x)$) while S-learners \textit{s}hare some information between nuisance estimators, which encompasses more strategies than just including the treatment indicator as a covariate (e.g. joint feature selection, or the strategies for NNs we discuss in  Section \ref{implementation}). 

We define two-step learners by their approach of using a first stage to obtain plug-in estimates $\hat{\eta}$ of (a subset of) the nuisance parameters $\eta=(\mu_0(x), \mu_1(x), \pi(x))$, and then a second stage to obtain an estimate $\hat{\tau}(x)$ by regressing a pseudo-outcome $\tilde{Y}_{\hat{\eta}}$ (based on nuisance estimates $\hat{\eta}$) on $X$ directly. To do so, we consider pseudo-outcomes for which it holds that $\mathbb{E}_{\mathbb{P}}[\tilde{Y}_{\eta}|X=x]=\tau(x)$ (i.e. they are unbiased for CATE when $\eta$ is known), for which there are three straightforward strategies inspired by ATE estimators:

(1) We propose an RA-learner, which uses the regression-adjusted pseudo-outcome 
\begin{equation}
\tilde{Y}_{RA, \hat{\eta}} = W(Y-\hat{\mu}_0(X)) + (1-W)(\hat{\mu}_1(X) - Y)
\end{equation}
in the second step.  \cite{kunzel2019metalearners}'s X-learner is a variant of this estimator: Instead of performing one regression in the second step, they suggest performing two separate regressions for each term in the sum, leading to two CATE estimators $\hat{\tau}_1(x)$ and $\hat{\tau}_0(x)$ that should then be combined into a final estimate using $\hat{\tau}(x) = g(x) \hat{\tau}_0(x) + (1 - g(x)) \hat{\tau}_1(x)$ for some weighting function $g(x)$. For $g(x)=1-\pi(x)$, the two estimators coincide in expectation. We prefer the RA-learning strategy here, as it does not require choice of `hyper-parameter' $g(x)$.

(2) Inspired by inverse propensity weighted (IPW) estimators, we consider a PW-learner with associated pseudo-outcome based on the Horvitz-Thompson transformation \citep{horvitz1952generalization}
\begin{equation}
\tilde{Y}_{PW, \hat{\eta}} = \left(\frac{W}{\hat{\pi}(X)}- \frac{1-W}{1-\hat{\pi}(X)}\right)Y 
\end{equation}

(3) Finally, the DR-learner \citep{kennedy2020optimal} has pseudo-outcome
\begin{equation}
\begin{split}
\tilde{Y}_{DR, \hat{\eta}} = \left(\frac{W}{\hat{\pi}(X)}- \frac{(1-W)}{1-\hat{\pi}(X)}\right) Y + \\ \left[\left(1 - \frac{W}{\hat{\pi}(X)}\right) \hat{\mu}_1(x)-\left(1 - \frac{1-W}{1-\hat{\pi}(X)}\right)\hat{\mu}_0(X)\right]
\end{split}
\end{equation}
which is based on the doubly-robust AIPW estimator \citep{robins1995semiparametric} and is hence unbiased if either propensity score \textit{or} outcome regressions are correctly specified. Variants of the DR-learner have previously also been studied in e.g. \cite{lee2017doubly, fan2020estimation}. 

The R-learner proposed in \cite{nie2017quasi} does not fall in any of these categories. While all learners considered above are inherently model-agnostic, i.e. can be implemented using any off-the-shelf ML method, the R-learner requires specific model-fitting procedures (e.g. the ability to manipulate a loss function).  We therefore do not consider this estimator further here, also because an in-depth theoretical analysis is given in \cite{nie2017quasi}.

\section{THEORETICAL ANALYSES OF CATE META-LEARNERS}
In this section we consider the theoretical behavior of the four different types of CATE meta-learners. While the asymptotic properties of both plug-in and DR-learner have been previously analyzed, our analyses of both RA- and PW-learner, as well as the comparison between all four strategies in asymptotic and finite sample settings, are new and are meant to provide insights to guide principled choice between algorithms. 

 Throughout, we denote by $\tau(x)$ the true CATE and by $\hat{\tau}_{l, \hat{\eta}}(x)$ the output of learner $l$. For two-step estimators we denote by $\hat{\tau}_{l, {\eta}}(x)$ the output of the second stage regression if we had oracle-access to the nuisance parameters. Further, for an estimator $\hat{f}(x)$ of $f(x)$, we denote by $\epsilon_{sq}(\hat{f}(x)) = \mathbb{E}[(\hat{f}(x)-f(x))^2]$ its expected squared error. Finally, we use $a\lesssim b$ to indicate $a \leq Cb$ for some universal constant $C$. For the purpose of the theoretical analysis in this section, we make the following two assumptions:

\begin{assumption}\label{esass}[Assumptions on estimators]
We assume for the propensity score estimates that $\delta \leq \hat{\pi}(x) \leq 1 - \delta$ for $\delta>0$. Further, all regression estimators fulfill the mild assumptions characterized in \cite{kennedy2020optimal}'s Theorem 1 (see supplement). 
\end{assumption}

\begin{assumption}\label{funcass}[Assumptions on true DGP]
We assume that $\mu_w(x)$ is $\alpha_w$-smooth, $\pi(x)$ is $\beta$-smooth and $\tau(x)$ is $\gamma$-smooth, and all are estimable at \cite{stone1980optimal}'s minimax rate of  $n^{\frac{-p}{2p+d}}$ for a $p$-smooth function (which has p continuous and bounded derivatives). Further, the potential outcomes and the propensity scores are bounded, i.e. $|\mu_w(x)| \leq C$ and $\omega \leq \pi(x)\leq 1 - \omega$ for $C, \omega>0$. 
\end{assumption}

\textbf{Asymptotic Behavior}
First, we consider the (asymptotic) behavior of $\epsilon_{sq}(\hat{\tau}_{l, \hat{\eta}}(x))=\mathbb{E}[(\hat{\tau}_{l, \hat{\eta}}(x) - \tau(x))^2]$. For a generic one-step plug-in estimator $\hat{\tau}(x) = \hat{\mu}_1(x) - \hat{\mu}_0(x)$ we almost trivially have that
\begin{equation*}
\epsilon_{sq}(\hat{\tau}(x)) \leq 2 [\epsilon_{sq}(\hat{\mu}_1(x)) + \epsilon_{sq}(\hat{\mu}_0(x))] \lesssim n^{\frac{-2\alpha_0}{2\alpha_0+d}} + n^{\frac{-2\alpha_1}{2\alpha_1+d}}
\end{equation*}
by $(a+b)^2 \leq 2 (a^2 + b^2)$ and assumption \ref{funcass}. The second inequality holds asymptotically for plug-in estimators based on T-strategies, but not necessarily for the S-learners because information-sharing might restrict the set of possible potential outcome functions.

The three other learners all require two regression stages and are hence more difficult to analyze. However, under the assumption that we perform the two regression stages on two separate samples $\mathcal{D}_0$ and $\mathcal{D}_1$ of size $n$, we can apply \cite{kennedy2020optimal}'s Theorem 1 , yielding that  $\epsilon_{sq}(\hat{\tau}_{l, \hat{\eta}}(x))$
\begin{equation*}
\lesssim \underbrace{\epsilon_{sq}(\hat{\tau}_{l,\eta}(x))}_{\lesssim n^{\frac{-2\gamma}{2\gamma+d}}} + \underbrace{\mathbb{E}[(\mathbb{E}[\tilde{Y}_{\hat{\eta}}(x)|X=x, \mathcal{D}_0] - \tau(x))^2]}_{ = R^2_{l, \hat{\eta}}(x)}
\end{equation*}
for all three learners $l$. As the oracle term is of fixed order across all estimators, the remainder term $R^2_{l, \hat{\eta}}(x)$, which we analyze in the following theorem (proof and additional analyses in supplement), will determine relative asymptotic performance. 

\begin{theorem}\label{mainthm}[Learner-specific remainders]
Under assumptions \ref{esass} and \ref{funcass} and using two-step estimation on two separate samples $\mathcal{D}_0$ and $\mathcal{D}_1$ of size $n$, we have that 
\\
(1) for the RA-learner: 
\begin{equation*}
\begin{split}
R^2_{RA, \hat{\eta}}(x) \leq 2(1-\omega)^2 \left(\sum_{w} \epsilon_{sq}(\hat{\mu}_w(x))\right) \\
\lesssim n^{\frac{-2\alpha_0}{2\alpha_0+d}} + n^{\frac{-2\alpha_1}{2\alpha_1+d}}
\end{split}
\end{equation*}

(2) for the PW-learner:
\begin{equation*}
R^2_{PW, \hat{\eta}}(x) \leq \frac{4C^2}{\delta^2} \epsilon_{sq}(\hat{\pi}(x))\lesssim  n^{\frac{-2\beta}{2\beta+d}}
\end{equation*}

(3) (\cite{kennedy2020optimal}, Theorem 2 and Corollary 1) If $\pi(x)$ and $\mu_w(x)$ are fit on separate sub-samples, we have for the DR-learner that:
\begin{equation*}
\begin{split}
R^2_{DR, \hat{\eta}}(x) \leq \frac{2}{\delta^2} \epsilon_{sq}(\hat{\pi}(x)) \left(\sum_{w} \epsilon_{sq}(\hat{\mu}_w(x))\right)
\\
\lesssim  n^{-2( \min_w \frac{\alpha_w}{2\alpha_w+d} + \frac{\beta}{2\beta+d})}
\end{split}
\end{equation*}
\end{theorem}

By asymptotic properties, we thus prefer the DR-learner over the other two-step learners, and the PW-learner over the RA-learner if $\beta > min_w \alpha_w$, i.e. if the propensity score is easier to estimate than the outcome regressions.  Further, if $\min_w \frac{\alpha_w}{2\alpha_w+d} + \frac{\beta}{2\beta+d} > \frac{\gamma}{2\gamma+d}$, the DR-learner attains the oracle rate, and so do PW- and RA-learner if $ \beta > \gamma$ and $min_w \alpha_w > \gamma$, respectively. The latter, however, is unlikely since it is commonly assumed that in practice $\tau(x)$ is simpler than $\mu_0(x)$ \citep{kunzel2019metalearners}. Asymptotically, we therefore expect RA- and plug-in learner to perform similarly. In the case where $\tau(x)$ is of similar complexity as the potential outcomes functions, plug-in and two-step learners can be expected to perform similarly. 
Instead of relying on assumed smoothness of the nuisance functions, similar analyses can be performed by relying on different assumptions on the problem structure. In the supplement, we consider sparsity instead of smoothness, which leads to analogous conclusions in terms of relative performance of the different learners. 

While in reality it is unlikely to have exact knowledge on the theoretical properties of the nuisance functions, the considerations above are also useful when an expert can give insight on the \textit{relative} complexity of the underlying functions. Further, in experimental settings, when propensity scores are \textit{known}, the remainder terms of PW- and DR-learner will be \textit{exactly} zero, rendering their performance equal to the oracle rate.

\textbf{Considerations for Finite Samples}
However, as we will discuss next, in smaller sample regimes, we cannot rely on convergence rates only. A first reason for this derives from the fact that while the oracle term $\mathbb{E}[(\hat{\tau}_{l,\eta}(x)- {\tau}(x))^2]$ is of the same order for all estimators, the variance of inverse propensity weighted estimators is well-known to be high, particularly when propensity scores are extreme. Additionally, the pseudo-outcome associated with the PW-learner has a very high variance even when propensity scores are constant and known, because it then still lies in $\{-2Y(0), 2Y(1)\}$. This -- as we will demonstrate in the experiments -- can lead to very poor empirical performance of the PW-learner. Similarly, for $\delta$ small, the constants in the error bounds of PW- and DR-learner in Theorem \ref{mainthm} can be large, possibly leading to relatively better performance of the RA-learner in smaller samples. 

\begin{figure*}[t]
\includegraphics[width=0.99\textwidth]{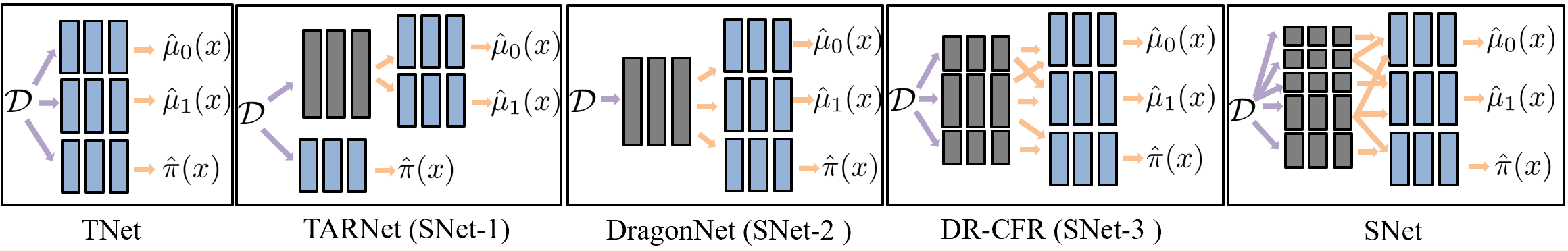}
\caption{Overview of five possible model architectures for one-step estimation of nuisance parameters, involving different levels of sharing information  between tasks (representation layers in gray, task-specific layers in blue)}\label{architecturefig}
\end{figure*}

Second, so far we did not consider \textit{selection bias} -- $\pi(x)\neq0.5$ for some $x$ -- which matters in finite sample regimes, as it results in each regression surface not being fit optimally. Consider the expected (integrated) error on the treated outcome surface, for which it is straightforward to show (see supplement) that:
\begin{gather}\label{eqselectionbias}
\textstyle \mathbb{E}_{X\sim\mathbb{P}(\cdot)}[(\hat{\mu}_1(X)-\mu_1(X))^2]=\\
 \textstyle \mathbb{E}_{X\sim\mathbb{P}(\cdot | W=1)}[w(X)(\hat{\mu}_1(X)-\mu_1(X))^2] \nonumber
\end{gather}
where $\textstyle{w(X)= p_\pi(1+\frac{1-\pi(X)}{\pi(X)})}$. When $\hat{\mu}_1(x)$ is fit on factual data only, this corresponds to setting $w(x)=1$ for all $x$, which gives too much weight to samples with large $\pi(x)$. Combining this with assumption \ref{funcass}, we have for one-step learners that $\mathbb{E}_{X\sim\mathbb{P}(\cdot)}[(\hat{\tau}(X)- \tau(X))^2]$
\begin{equation*}
\begin{split}\leq \frac{2}{\omega}\big(\mathbb{E}_{X\sim\mathbb{P}(\cdot | W=1)}[(\hat{\mu}_1(X) - \mu_1(X))^2]  \\ + \mathbb{E}_{X\sim\mathbb{P}(\cdot | W=0)}(\hat{\mu}_0(X) - \mu_0(X))^2]\big) 
\end{split}
\end{equation*}
which highlights why selection bias does not matter asymptotically, but can be severe in finite samples if $\omega$ is small (a similar conclusion was reached in \cite{alaa2018limits}). It also shows that the degree of overlap, which determines $\omega$, is of paramount importance. Because the second stage regressions use all data to estimate $\tau(x)$ directly, and hence consider $X \sim \mathbb{P}(\cdot)$, this can correct for sub-optimal weighting in the first stage.

Overall, we conclude that a second stage regression can act like a `regularizer' on the first stage output, removing some of the bias induced by regularization and overfitting in finite samples. We note that the DR-learner does so optimally from a theoretical viewpoint, as it can also be shown to build on the notion of nonparametric plug-in bias removal via influence functions \citep{curth2020semiparametric}. Nonetheless, we note that two-step learners generally need to estimate more parameters on the same amount of data (or, if sample splitting is used, on less), which could in practice also lead to higher variance and worse performance in low sample regimes.

\section{CATE ESTIMATION USING NEURAL NETWORKS}\label{implementation}
In the practical part of this paper, we use feed-forward NNs as nuisance estimators for each meta-learner. The simplest NN-based implementation of each learner consists of using a separate network for each regression task (a TNet), which would be a good choice asymptotically, allowing for arbitrarily different regression surfaces. 

However, as we alluded to in Section 3, it can be useful to share information between nuisance estimation tasks in finite samples. 
That is, it may be more \textit{efficient} to share data between the two regression tasks if $\mu_1(x)$ and $\mu_0(x)$ are similar, which would be the case if they were supported on similar covariates and $\tau(x)$ is not too complex. Therefore, next to a simple T-learner (TNet), we consider a class of model architectures we refer to as SNets because they are based on \textit{s}haring information between nuisance estimation tasks using representation learning. The resulting one-step architectures for nuisance estimation are visualized in Fig. \ref{architecturefig}, and we discuss implementation details and loss-functions in the supplement.

\textbf{Existing SNet Architectures}
Building on the success of representation learning on a variety of learning tasks \citep{bengio2013representation}, \cite{shalit2017estimating} introduce the idea of learning a shared input representation for the two potential outcome regressions. Formally, this entails jointly learning a map $\Phi: \mathcal{X} \rightarrow \mathcal{R}$, representing \textit{all} data in a new space, and two regression heads $\mu_w: \mathcal{R} \rightarrow \mathcal{Y}$, fit using only the data of the corresponding treatment group. This results in TARNet \citep{shalit2017estimating}, which we will also refer to as SNet-1 because it results in the simplest way of sharing information between tasks.  \cite{shi2019adapting}'s DragonNet (SNet-2) takes this idea one step further and learns a representation space from which both $\pi(x)$ and $\mu_w(x)$ can be learned, ensuring that confounders are sufficiently controlled for. While \cite{shi2019adapting} used this architecture, combined with ideas from the targeted maximum likelihood estimation framework,  to estimate \textit{average} treatment effects, we use only their model architecture for CATE estimation. Finally, \cite{hassanpour2019learning}'s DR-CFR (SNet-3) learns three representations which are used to model \textit{either} propensity score, potential outcome regressions or both, respectively\footnote{To limit the number of possible models, we consider only straightforward plug-in strategies, and do not use the propensity heads for re-weighting within the loss function (as \cite{hassanpour2019learning} do).}. 

\textbf{Underlying Assumptions and a New Architecture}
Each existing architecture reflects implicit assumptions on the underlying structure of the CATE estimation problem. Therefore, explicitly characterizing assumptions can help choosing between architectures in practice, and guide improvements by identifying shortcomings. SNet-1 builds on the assumption that there exists a common feature space underlying both $\{\mu_w(x)\}_{w\in\{0,1\}}$, while SNet-2 assumes that $\pi(x)$ can \textit{also} be represented in the same space. SNet-3 is built on the assumption that there exist three separate sets of features, determining $\mu_w(x)$  and/or $\pi(x)$.

Reflecting on these assumptions, we note that there is one important case missing: We wish to explicitly allow for the existence of features that affect \textit{only one} of the potential outcome functions. This is driven by  medical applications where one distinguishes between markers that are prognostic (of outcome) regardless of treatment status or predictive (of treatment effectiveness) \citep{ballman2015biomarker}. Therefore, we propose a final model architecture that learns five representations\footnote{Note that, when multiple representations and outcome functions are learned jointly, the distinction between representations is not necessarily well-identified. Therefore, for both SNet-3 and SNet we use a regularization term inspired by \cite{wu2020learning} that enforces orthogonalization of inputs to the different representation layers, as we discuss in the supplement.}, also allowing potential-outcome regressions to depend on only a subset of shared features. We will simply refer to this architecture as SNet because it \textit{encompasses} all existing SNet-architectures and even the TNet as special cases. That is, if we increase the width of the $\mu_w(x)$-specific representations while reducing the shared representations, SNet will approach TNet. If we do the reverse, the proposed architecture approaches SNet-3, which can in turn become either SNet-1 or SNet-2 by changing what is shared between $\mu_w(x)$ and $\pi(x)$. We expect that such a general architecture should perform best \textit{on average} in absence of knowledge on the underlying problem structure, but may be more difficult to fit than simpler models.

\textbf{Further Practical Considerations} Two-step learners are most commonly implemented using independently trained ``vanilla'' nuisance estimators, yet, as we investigate in the experiments, all SNet architectures could also be used to estimate nuisance parameters in the first step. Further, while our theoretical analyses rely on sample splitting for two-step learners, we observed that using all data for both steps can work better in practice (particularly in small samples). If estimation with theoretical guarantees is desired yet data is scarce, it can be useful to rely not on sample splitting, but on \textit{cross-fitting} \citep{chernozhukov2018double} to obtain valid nuisance function estimates for all observations in the sample, which can then all be used for a second stage regression. Which strategy to use involves a trade-off between precision in estimation and computational complexity. While we implement all strategies in our code base, we do not use any form of sample splitting in our experiments. Finally, a convenient by-product of splitting the causal inference task into a series of multiple supervised learning tasks is that hyper-parameters can be tuned using factual hold-out data only -- which can be done in both regression stages, if data is split appropriately.

\section{EXPERIMENTS}
In this section we supplement our theoretical analyses with experimental evidence to demonstrate the empirical performance of all learners under different DGPs, using both fully synthetic data and the well-known semi-synthetic IHDP benchmark. In addition to verifying the theoretical properties of the different learners empirically, we consider it of particular interest to gain insight into how to best use the different NN architectures as nuisance estimators for the two-step learners. 

Throughout the experiments, we fixed equivalent hyper-parameters across all model architectures (based on those used in \cite{shalit2017estimating}), ensuring that every estimator (output head) has access to the same total amount of hidden layers and units, and effectively used each learner `off-the-shelf'.   To ensure fair comparison across learners, we implemented every architecture in our own python code base\footnote{All code is available at \url{https://github.com/AliciaCurth/CATENets} and at \url{https://github.com/vanderschaarlab/mlforhealthlabpub/tree/main/alg/CATENets}}; for implementation details, refer to the supplement. Throughout, we consider performance in terms of the Root Mean Squared Error (RMSE) of estimation of $\tau(x)$, also sometimes referred to as the precision of estimating heterogeneous effects (PEHE) criterion \citep{Hill2011}.  

\subsection{Synthetic Experiments}
We simulate data to investigate the relative performance of the different learners across different underlying DGPs and sample sizes. Throughout, we consider $d=25$ multivariate normal covariates, of which we let subsets determine $\mu_w(x)$ and $\pi(x)$. To highlight scenarios under which different learners can be expected to perform well,  we consider three (highly stylized) settings: (i) $\tau(x)=0$, and $\mu_0(x)$ depends on 5 covariates influencing outcome and 5 confounders (which influence also $\pi(x)$), (ii) the same set-up as (i), but with $\tau(x)$ nonzero and supported on 5 additional covariates and (iii) $\mu_1(x)$ and $\mu_0(x)$ depend on disjoint covariate sets, making $\tau(x)$ the most difficult function to estimate (no confounders). All DGPs are discussed in detail in the supplement. In all simulations, we evaluate performance on 500 independently generated test-observations, and average across 10 runs. 

\begin{figure}[t]
\includegraphics[width=\columnwidth]{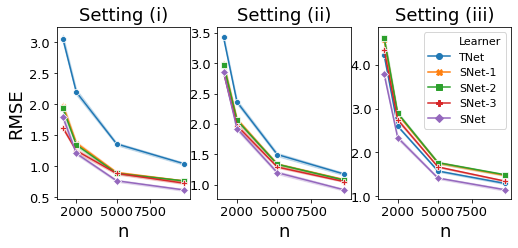}
\caption{RMSE of plug-in architectures by sample size and across different DGPs. Shaded area indicates one standard error.}\label{pluginres}
\end{figure}

When comparing the 5 plug-in architectures (Fig. \ref{pluginres}), we note that S-architectures always improve upon the TNet in small samples when $\{\mu_w\}_{w \in \{0,1\}}$ share some structure. Even when $\mu_1(x)$ and $\mu_0(x)$ are very different, shared layers can help by filtering out noise covariates. Further, the flexibility of the respective SNet architecture seems to indeed drive performance. As expected, the general SNet performs best on average, and is the only architecture to outperform TNet when $\mu_1(x)$ and $\mu_0(x)$ are very different. SNet performs well also in the absence of any treatment effect, which we attribute to the architecture's ability to learn that there are no predictive features to represent using the $\mu_w(x)$-specific representations. Further, as expected, the strength of SNet relative to all variants becomes most apparent in relatively larger sample sizes. 

\begin{figure}[b]
\includegraphics[width=\columnwidth]{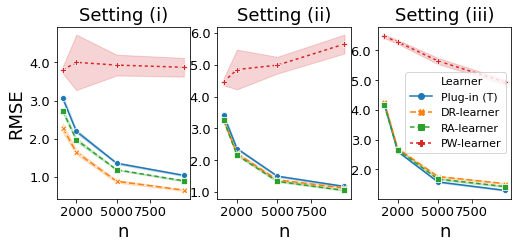}
\caption{RMSE of different meta-learners by sample size and across different DGPs. Shaded area indicates one standard error.}\label{strategyres}
\end{figure}

\begin{figure}[t]
\includegraphics[width=\columnwidth]{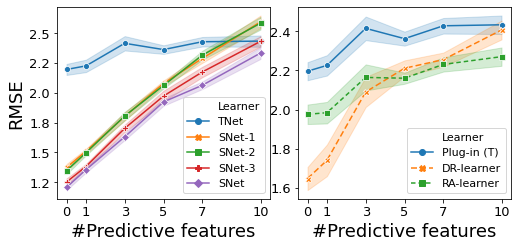}
\caption{RMSE of different plug-in architectures (left) and  meta-learners (right) by number of predictive features at $n=2000$. Shaded area indicates one standard error.}\label{predres}
\end{figure}

When comparing the four learning strategies (Fig. \ref{strategyres}), all based on TNet in the first step, we observe that the DR-learner substantially outperforms the others when there is confounding but no treatment effect (setting (i)). Conversely, when CATE is \textit{more} complex than either potential outcome function (setting (iii)) -- making the task of learning CATE directly more difficult than learning the potential outcome functions separately -- the plug-in learner performs best, as expected. The RA-learner performed best when there is both confounding and a non-trivial treatment effect (setting (ii)). The PW-learner performed poorly in general,  which is caused by the very low signal-to-noise ratio and high variance in the associated pseudo-outcome. While all other learning strategies performed very well using equivalent architectures, optimizing the PW-learner would require much stronger regularization and less flexibility (smaller networks) than what we considered here.

To gain further insight to the relative performance of different learning strategies, we interpolate between settings (i) and (ii) by gradually increasing the number of predictive features (features determining $\tau(x)$) at $n=2000$ in Fig. \ref{predres}. We observe that the performance of the different S-architectures degrades relative to TNet as the number of predictive features increases, which is to be expected as $\mu_0(x)$ and $\mu_1(x)$ become less similar. We also observe that while the performance gap between TNet and RA-learner remains virtually constant, the DR-learner loses its advantage as CATE becomes less sparse.

\begin{figure}[b]
\includegraphics[width=\columnwidth]{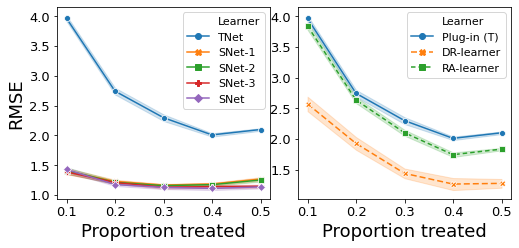}
\caption{RMSE of different plug-in architectures (left) and  meta-learners (right) by proportion of treated individuals in setting (i) at $n=2000$. Shaded area indicates one standard error.}\label{propres}
\end{figure}

In Fig. \ref{propres} we reconsider setting (i) but with \textit{imbalance} in addition to confounding, by gradually changing the proportion of treated individuals. Comparing the different plug-in architectures, we observe that information sharing has a much larger added value when samples are highly imbalanced. Comparing the different learners, we observe that the performance gap between T- and DR-learner is not impacted by the proportion of treated individuals, but that the RA-learner outperforms the T-learner only for moderate to no imbalance.

\begin{figure}[t]
\includegraphics[width=\columnwidth]{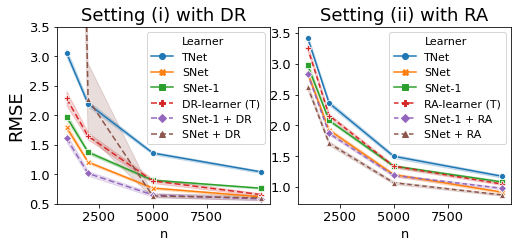}
\caption{RMSE of different learner-architecture combinations by sample size and across different DGPs. Shaded area indicates one standard error.}\label{combores}
\end{figure}

Finally, we consider how to best combine nuisance estimators and two-step learners (Fig. \ref{combores}). We reconsider settings (i) and (ii) where DR- and RA-learner performed best, and use SNet-1, which does not estimate $\pi(x)$, and the more general SNet, which does, as nuisance estimators. For the RA-learner, we observe that using SNet, which has better performance on its own, leads to the best RA-learner and conclude that in practice the RA-learner should be combined with an architecture that is expected to best capture the underlying DGP. For the DR-learner, we note that strong dependence between $\hat{\pi}(x)$ and $\hat{\mu}_w(x)$ leads to a slower decaying remainder term theoretically, manifesting in poor empirical performance of using SNet with the DR-learner relative to SNet-1 in smaller samples. Hence, the DR-learner is best combined with a nuisance estimator that does not share information between estimation of $\pi(x)$ and $\mu_w(x)$.

\subsection{Semi-synthetic Benchmark: IHDP}
Additionally, we deploy all algorithms on the well-known IHDP benchmark, based on real covariates with simulated outcomes. We use an adapted\footnote{We noticed that the scale of CATE varies by orders of magnitude across different settings in the original data, making (R)MSE incomparable across runs. As discussed in the supplement, we rescaled the responses to correct for this.} version of the 100 realizations provided by \cite{shalit2017estimating}. The data-set is small ($n=747$, of which 90\% are used in training),  imbalanced (19\% treated), and there is only partial overlap \citep{Hill2011}. While the two potential outcome functions are supported on the same covariates, their functional forms are different, making their difference -- CATE -- the most difficult function to estimate. A more detailed description of the data-set can be found in the supplement.

In Table \ref{resultsihdp}, we observe that information sharing in plug-in learners indeed significantly improves performance versus the simple TNet, and that more complex models (SNet-3 \& SNet) underperform simpler models (SNet-1 \& 2). Both of these observations are not surprising given the small sample size, the high treatment group imbalance, as well as the fact that there is no true separation of covariates into different adjustment sets in the DGP. Further, we observe that -- with the exception of using the RA-Learner on top of the best-performing plug-in model -- the two-step learners underperform the plug-in learners on the IHDP data-set, a direct consequence of the complexity of the simulated $\tau(x)$. Further, potentially because overlap is incomplete in this data-set, the RA-learner outperforms the DR-learner substantially.
\begin{table}[t]
\small
\caption{RMSE of all learners on the adapted IHDP data-set. Averaged across 100 realizations, standard error in parentheses.}\label{resultsihdp}
\begin{tabular}{lll}
\toprule
Model               & In-Sample     & Hold-out      \\ \midrule
TNet                & 0.761 (0.011) & 0.770 (0.013) \\
SNet-1              & 0.678 (0.009) & 0.689 (0.012) \\
SNet-2              & 0.676 (0.009) & 0.687 (0.012) \\
SNet-3              & 0.683 (0.009) & 0.691 (0.011) \\
SNet             & 0.729 (0.009) & 0.737 (0.011) \\
RA-Learner + TNet   & 0.740 (0.010) & 0.745 (0.013) \\
RA-Learner + SNet-2 & \textbf{0.670} (0.009) & \textbf{0.680} (0.012) \\
DR-Learner + TNet   & 0.893 (0.017) & 0.902 (0.019) \\
PW-Learner + TNet        & 2.250 (0.093) & 2.285 (0.094)\\ \bottomrule
\end{tabular}
\end{table} 

\section{CONCLUSION}
In this paper we considered meta-learning strategies for nonparametric CATE estimation, theoretically analyzed their properties, and implemented them using a range of neural network architectures. We demonstrated that while the DR-learner is asymptotically optimal in theory, both the RA-learner and plug-in learners sharing information between nuisance estimation tasks can outperform it in finite samples. We also showed that using sophisticated architectures as nuisance estimators for two-step learners instead of vanilla NNs can boost their small sample performance. In addition, we highlighted that the relative performance of different learners and architectures depends both on the underlying DGP and the amount of data at hand, such that the choice of learner in practice should incorporate an expert's assessment of the most likely DGP. While we considered only the choice between meta-learners using the same underlying method throughout, investigating the optimal choice of ML method -- e.g. NNs versus random forests -- would be an interesting next step.
\newpage
\subsubsection*{Acknowledgements} We thank anonymous reviewers as well as members of the vanderschaar-lab for many insightful comments and suggestions. AC gratefully acknowledges funding from AstraZeneca. 

\bibliography{representation}
\bibliographystyle{apalike}

\appendix
\onecolumn
\section*{SUPPLEMENTARY MATERIALS}
\section{ASSUMPTIONS AND ADDITIONAL ANALYSES}
In this section, we revisit the assumptions made in Section 4. First, we discuss estimation under assumed sparsity instead of smoothness, and then we discuss the assumptions associated with \cite{kennedy2020optimal}'s theorem on pseudo-outcome regression. 
\subsection{Additional Analyses on Minimax Performance Using Assumptions on Sparsity}
\paragraph{Why consider minimax error rates?} In the main text, we relied on assumptions of smoothness on underlying functions and below we discuss estimation under assumed sparsity. We do so to nonparametrically \textit{quantify} the hardness of the different estimation problems, allowing us to systematically compare the minimax performance of different learners. The remainders derived in Theorem 1 allow for much more general analyses than what we discuss here, and could be used to assess the relative performance of the different learners using \textit{any} assumption on the learning rates associated with the ``difficulty'' of the functions $\tau(x)$, $\mu_w(x)$ and $\pi(x)$. We rely on smoothness and sparsity due to their intuitive appeal, generality and usage in related work \citep{alaa2018limits, kennedy2020optimal}.
\paragraph{Minimax rates for estimation under sparsity} Previously, we relied on assumed smoothness of the different regression functions, to illustrate the effect of differences in underlying complexity of the nuisance functions. Instead of smoothness only, we now consider functions with sparsity and additive sparsity as defined in assumption M3 in \cite{yang2015minimax}, which is often a necessary assumption enabling estimation when data is high-dimensional (e.g. $d>n$, where $X \in \mathbb{R}^d$). A function $f$ satisfies (additive) sparsity if it depends on $d^* \asymp min(n^\gamma, d)$ variables for some $\gamma \in (0, 1)$ but admits an additive structure $f = \sum^k_{s=1} f_s$ where each of the $k$ component functions $f_s$ depends on a small number of predictors ($d_s$). As special cases of this assumption we have both the more standard sparsity assumption, where one $f$ depends on one small subset $d^* \leq min(n, d)$ of the predictors (i.e. $k=1$), as well as the case where $f$ is completely additive ($d_s=1$ for all $s$).

 For ease of exposure, we also assume that all additive components $f_s$ have the same smoothness $p_s=p$, dimension $d_s=d^*$ and magnitude. As shown by \cite{yang2015minimax}, this leads to the minimax rate $kn^{-2p/(2p + d^*)} + k\frac{d^* \log(d/d^*)}{n}$ in squared error of estimation of $f$. The first term is \cite{stone1980optimal}'s nonparametric minimax rate for a p-smooth function as we considered in the main text, but with $d^* \leq d$ known, while the second term captures the uncertainty in variable selection.
 \paragraph{Learner performance under sparsity}  We can use this minimax rate to compare the error rates attained by the different learners similarly as we used smoothness in the main text. For example, if we assume that each regression function $f$ is $d^*$-sparse and linear in $X$, we have the minimax rate $\frac{d^* \log(d/d^*)}{n}$ for an estimator $\hat{f}$ \citep{raskutti2009lower}, and the squared error of estimation using the lasso can attain the minimax rate $\frac{d^*log(d)}{n}$ \citep{bickel2009simultaneous}. 
 
 \begin{assumption}
 Assume that $\tau(x)$, $\mu_w(x)$ and $\pi(x)$ are linear in $x$ and $d_\tau$, $d_{\mu_w}$ and $d_\pi$-sparse, respectively, and each function $f$ can hence be estimated with squared error rate $\frac{d_f log(d)}{n}$. 
 \end{assumption}

 Then, we immediately have the following corollary:

 \begin{corollary} Using Theorem 1, we have the following error rates on the four learners under the sparsity assumptions discussed above:
 \begin{itemize}
     \item For the plug-in learner:
     \begin{equation}
        \mathbb{E}[(\hat{\tau}_{plug, \hat{\eta}}(x) - \tau(x))^2] \lesssim \frac{(d_{\mu_0} + d_{\mu_1}) log(d)}{n}
     \end{equation}
     \item For the RA-learner:
     \begin{equation}
      \mathbb{E}[(\hat{\tau}_{RA, \hat{\eta}}(x) - \tau(x))^2]\lesssim \frac{d_\tau log(d)}{n} + \frac{(d_{\mu_0} + d_{\mu_1}) log(d)}{n}
      \end{equation}
     \item For the PW-learner:
    \begin{equation}
      \mathbb{E}[(\hat{\tau}_{PW, \hat{\eta}}(x) - \tau(x))^2]\lesssim  \frac{d_\tau log(d)}{n} + \frac{d_\pi log(d)}{n}
      \end{equation}
     \item For the DR-learner:
    \begin{equation}
      \mathbb{E}[(\hat{\tau}_{DR, \hat{\eta}}(x) - \tau(x))^2]\lesssim \frac{d_\tau log(d)}{n} + \frac{(d_{\mu_0} + d_{\mu_1})d_\pi log^2(d)}{n^2}
      \end{equation}
 \end{itemize}
 \end{corollary}
This leads to analogous conclusions on learner performance as presented in the main text, but based on sparsity instead of smoothness. For example, two-step learners can outperform the plug-in learner if $d_\tau < max(d_{\mu_1}, d_{\mu_0})$, i.e. if the treatment effect depends on less covariates than each potential outcome function. This would \textit{not} be the case if $\mu_1(x)$ and $\mu_0(x)$ would depend on completely disjoint sets of covariates, as in setting (iii) in our experiments. Further, if  $d_\tau < max(d_{\mu_1}, d_{\mu_0})$, then RA-learner and plug-in learner are expected to perform equally, and PW- outperforms RA-learner if $d_\pi < max(d_{\mu_1}, d_{\mu_0})$. Finally, the DR-learner attains the oracle rate for CATE estimation if $d_\tau > \frac{(d_{\mu_0} + d_{\mu_1})d_\pi log(d)}{n}$ (this is shown also in \cite{kennedy2020optimal}).

Using the more general formulation of \cite{yang2015minimax}, relying not on linear but general sparse functions, would allow to consider even more general scenarios and derive conditions under which each learner outperforms the others based on smoothness of nonlinear functions \textit{and} sparsity. 


\paragraph{Further comparison with existing results} As previously mentioned, some of the different learners have been discussed separately in related work. In particular, asymptotic analyses for the DR-learner under smoothness and sparsity assumptions were presented in \cite{kennedy2020optimal}, and \cite{alaa2018limits} derive an error bound for Bayesian estimation of treatment effects using plug-in learners, based on an information-theoretic approach assuming smoothness and sparsity, which gives results for the plug-in learner analogous to ours. Yet, by analyzing the four main strategies for nonparametric meta-learning of CATE within one coherent framework, we contribute to existing work by building systematic understanding of the relative strengths and weaknesses of different estimation strategies in different scenarios. In particular, theoretical comparisons between one- and two-step learners often emphasize the favorable properties of two-step learners (e.g. for the X-learner in \cite{kunzel2019metalearners} and the DR-learner in \cite{kennedy2020optimal}) because it is typically assumed that CATE is much simpler than the baseline outcome function $\mu_0(x)$ \citep{kunzel2019metalearners}. This assumption, however, is not always reflected in the DGPs used to evaluate performance of CATE estimators elsewhere. The IHDP benchmark used in (among others) \cite{shalit2017estimating, shi2019adapting, hassanpour2019learning, wu2020learning}, based on \cite{Hill2011}'s simulation setting B (discussed in detail below), does \textit{not} satisfy this assumption, leading to conclusions based on empirical performance that seemingly stand in conflict with theoretical analyses highlighting mainly the favorable properties of two-step learners.

\subsection{Assumptions on Estimators for Pseudo-outcome Regression}
 To be able to bound the error in pseudo-outcome regression using \cite{kennedy2020optimal}'s Theorem 1 (which leads to the additive error decomposition using the oracle rate of estimation of CATE) we need a mild assumption on the second-stage regression model necessary to ensure stability of the second stage regression \citep{kennedy2020optimal}: 
\begin{assumption}\label{kennedyass} \textbf{Regularity of regression estimators}\\
We need two mild assumptions on the regularity of our second-stage regression estimators $\hat{\mathbb{E}}_n$. $\hat{\mathbb{E}}_n$ needs to satisfy that:
\begin{enumerate}
\item $\hat{\mathbb{E}}_n(Y|X=x) + c = \hat{\mathbb{E}}_n(Y+c|X=x)$ for any constant c
\item If $\mathbb{E}[Y|X=x]= \mathbb{E}[W|X=x]$ then 
\begin{equation*}
	\mathbb{E}\left[\{\hat{\mathbb{E}}_n[W|X=x] - \mathbb{E}[W|X=x]\}^2\right] \asymp\mathbb{E}\left[\{\hat{\mathbb{E}}_n[Y|X=x] - \mathbb{E}[Y|X=x]\}^2\right]
\end{equation*}
\end{enumerate}
\end{assumption}
The first assumption enforces that adding a constant to an outcome pre- or post-regression leads to the same result, whereas the second assumption says that the regression method results in the same error (up to constants) for two variables with the same conditional means, regardless of e.g. variance \citep{kennedy2020optimal}. 

\section{PROOFS}
\subsection{Proof of Theorem 1}
Here, we consider the term $\mathbb{E}[(\mathbb{E}[\tilde{Y}_{\hat{\eta}}(x)|X=x, \mathcal{D}_0] - \tau(x))^2]$ in detail for each learner. For the DR-learner, this was proven in \cite{kennedy2020optimal}, but we restate the proof here for completeness. By the tower property, we have for the term $R = \mathbb{E}[\tilde{Y}_{\hat{\eta}}(x)|X=x, \mathcal{D}_0] - \tau(x)$ for each two-step learner:\\
(1) RA-Learner: 
\begin{equation*}
\begin{split}
R = \pi(x)[\mu_1(x) - \hat{\mu}_0(x)] + (1-\pi(x))[\hat{\mu}_1(x) - \mu_0(x)] - (\mu_1(x) - \mu_0(x)) \\ = \pi(x)[\mu_0(x) - \hat{\mu}_0(x)] + (1-\pi(x))[\hat{\mu}_1(x) -\mu_1(x)]
\end{split}
\end{equation*}
(2) PW-Learner:
\begin{equation*}
\begin{split}
R = \frac{\pi(x)}{\hat{\pi}(x)}\mu_1(x) - \frac{1- \pi(x)}{1- \hat{\pi}(x)}\mu_0(x) - (\mu_1(x) - \mu_0(x)) \\ =
(\frac{\pi(x)}{\hat{\pi}(x)}-1)\mu_1(x) - (\frac{1- \pi(x)}{1- \hat{\pi}(x)}-1)\mu_0(x)\\
= \frac{1}{\hat{\pi}(x)}(\pi(x) - \hat{\pi}(x))\mu_1(x) - \frac{1}{1- \hat{\pi}(x)}(\hat{\pi}(x)-\pi(x))\mu_0(x)
\end{split}
\end{equation*}
(3) DR-Learner:
\begin{equation*}
\begin{split}
R = \frac{1}{\hat{\pi}(x)}(\pi(x) - \hat{\pi}(x))(\hat{\mu}_1(x)  - \mu_1(x)) - \frac{1}{1- \hat{\pi}(x)}(\hat{\pi}(x)-\pi(x))(\hat{\mu}_0(x) - \mu_0(x))
\end{split}
\end{equation*}

Using the identity $(a+b)^2 \leq 2 (a^2 + b^2)$ and assumption 2, this yields for the square $R^2 = (\mathbb{E}[\tilde{Y}_{\hat{\eta}}(x)|X=x, \mathcal{D}_0] - \tau(x))^2$ for the \\
(1) RA-learner:
\begin{equation*}
\begin{split}
R^2 \leq 2(\pi(x)^2[\mu_0(x) - \hat{\mu}_0(x)]^2 + (1-\pi(x))^2[\hat{\mu}_1(x) -\mu_1(x)]^2) \\
\leq 2(1-\omega)^2(\sum_{w \in \{0, 1\}} (\hat{\mu}_w(x) - \mu_w(x))^2)
\end{split}
\end{equation*}
(2) PW-learner:
\begin{equation*}
\begin{split}
R^2 \leq 2 (\frac{\mu^2_1(x)}{\hat{\pi}^2(x)} + \frac{\mu^2_0(x)}{(1-\hat{\pi}(x))^2})(\hat{\pi}(x) - \pi(x))^2 
\leq \frac{4C^2}{\delta^2} (\hat{\pi}(x) - \pi(x))^2
\end{split}
\end{equation*}
(3) DR-learner:
\begin{equation*}
\begin{split}
R^2 \leq 2 (\hat{\pi}(x) - \pi(x))^2 (\frac{(\hat{\mu}_1(x) - \mu_1(x))^2}{\hat{\pi}^2(x)} + \frac{(\hat{\mu}_0(x) - \mu_0(x))^2}{(1 - \hat{\pi})^2}) \\
\leq \frac{2}{\delta^2} (\hat{\pi}(x) - \pi(x))^2 (\sum_{w \in \{0, 1\}} (\hat{\mu}_w(x) - \mu_w(x))^2)
\end{split}
\end{equation*}

The theorem follows by taking expectations over $R^2$, and applying assumption 3. 

\subsection{Proof of equation 5}
In the following, we denote by $p(\cdot)$ the pdf of $\mathbb{P}$. Further, we let $R_1= \mathbb{E}_{X\sim\mathbb{P}(\cdot | W=1)}[(\hat{\mu}_1(X)-\mu_1(X))^2]$. By repeated application of Bayes rule and the law of total probability we can show that: 
\begin{equation*}
\mathbb{E}_{X\sim\mathbb{P}(\cdot)}[(\hat{\mu}_1(x)-\mu_1(x))^2] = \int (\hat{\mu}_1(x)-\mu_1(x))^2 p(x) dx 
\end{equation*}
\begin{equation*}
= \mathbb{P}(W=1) \int (\hat{\mu}_1(x)-\mu_1(x))^2 p(x|W=1) dx + (1-\mathbb{P}(W=1)) \int (\hat{\mu}_1(x)-\mu_1(x))^2 p(x|W=0) dx 
\end{equation*}
\begin{equation*}
= \mathbb{P}(W=1) R_1 + (1-\mathbb{P}(W=1)) \int (\hat{\mu}_1(x)-\mu_1(x))^2 \frac{p(x|W=1)}{p(x|W=1)} p(x|W=0) dx 
\end{equation*}
\begin{equation*}
= \mathbb{P}(W=1) R_1 + (1-\mathbb{P}(W=1)) \int (\hat{\mu}_1(x)-\mu_1(x))^2 \frac{\frac{\mathbb{P}(W=0|x)p(x)}{(1-\mathbb{P}(W=1))}}{\frac{\mathbb{P}(W=1|x)p(x)}{\mathbb{P}(W=1)}} p(x|W=1) dx 
\end{equation*}
\begin{equation*}
= \mathbb{P}(W=1) R_1 + \mathbb{P}(W=1) \int (\hat{\mu}_1(x)-\mu_1(x))^2 \frac{\mathbb{P}(W=0|x)}{\mathbb{P}(W=1|x)} p(x|W=1) dx 
\end{equation*}
\begin{equation*}
= \mathbb{P}(W=1) \int \left(1 + \frac{\mathbb{P}(W=0|x)}{\mathbb{P}(W=1|x)}\right) (\hat{\mu}_1(x)-\mu_1(x))^2  p(x|W=1) dx 
\end{equation*}
\begin{equation*}
=\mathbb{E}_{X\sim\mathbb{P}(\cdot | W=1)}\left[\mathbb{P}(W=1)\left(1+\frac{1-\pi(X)}{\pi(X)}\right)(\hat{\mu}_1(X)-\mu_1(X))^2\right]
\end{equation*}

\section{LEARNING ALGORITHMS AND IMPLEMENTATION}
In this section we first give pseudo code for the two-step learners, then discuss the loss functions associated with the different SNets, and finally discuss implementation details. 
\subsection{Two-step learner pseudo code}
Below, we present the pseudo code for the two-step learners. As discussed in Section 5, we used no form of sample splitting (option 3 in the algorithm described below) in our experiments, yet both cross-fitting (option 1) and a single sample split (option 2) could be used to implement a two-step learner for which the theoretical guarantees hold as analysed.
\algcaption{Two-step learner}
\begin{algorithmic}[1]
\STATE	\textbf{Inputs}: A sample $\mathcal{D}=\{Y_i, W_i, X_i\}^n_{i=1}$, a learning algorithm $\mathcal{A}$, a first-stage fitting strategy and a second stage pseudo-outcome formula $\tilde{Y}_{\hat{\eta}}=f_{\tilde{Y}}(Y, W, X; \hat{\eta})$
\STATE \textbf{First stage: nuisance model estimation}
\IF{fitting strategy is cross-fitting}
\STATE split the sample $\mathcal{D}$ in $k$ non-overlapping folds
\FOR{$k \leftarrow 1:K$}
\STATE Fit nuisance models $\hat{\eta}_{-k}=\mathcal{A}(\mathcal{D}_{-k})$ on all but the $k^{th}$ fold
\STATE Predict $\tilde{Y}_i = f_{\tilde{Y}}(Y_i, W_i, X_i; \hat{\eta}_{-k})$ for $i \in \mathcal{D}_k$ using the nuisance model $\hat{\eta}_{-k}$
\ENDFOR 
\ELSIF{Fitting strategy is sample splitting}
\STATE split the sample into $\mathcal{D}_1$ and $\mathcal{D}_2$
\STATE Fit nuisance model $\hat{\eta}=\mathcal{A}(\mathcal{D}_{1})$ on $\mathcal{D}_1$
\STATE Predict $\tilde{Y}_i = f_{\tilde{Y}}(Y_i, W_i, X_i; \hat{\eta})$ for $i \in \mathcal{D}_2$ using the nuisance model $\hat{\eta}$
\ELSE
\STATE Fit nuisance model $\hat{\eta}=\mathcal{A}(\mathcal{D})$ on full sample
\STATE Predict $\tilde{Y}_i = f_{\tilde{Y}}(Y_i, W_i, X_i; \hat{\eta})$ for $i \in \mathcal{D}$ using the nuisance model $\hat{\eta}$
\ENDIF
\STATE \textbf{Second stage: CATE estimation}
\STATE estimate $\tau(x)$ as a function of $x$ by regressing $\{\tilde{Y}_i\}$ on $\{X_i\}$ as $\hat{\tau}(x) = \mathcal{A}(\{\tilde{Y}_i, X_i\})$
\STATE \textbf{Output}: $\hat{\tau}(x)$
\end{algorithmic} 

\subsection{Loss functions for SNets}
In this section we present the loss functions we use to implement all SNet variants. For SNets 1 - 3, these are adapted from \cite{shalit2017estimating}, \cite{shi2019adapting} and \cite{hassanpour2019learning}, respectively, but not exactly identical -- we did not use re-weighting, re-balancing or TMLE-regularization schemes in estimating nuisance parameters, as we wish to consider only direct plug-in estimators. Strictly speaking, we therefore adapted only their model architectures. Further, it would be possible to assign different weights to different loss components (e.g. loss on propensity score estimation) below, however, we do not do so here to avoid adding additional hyper-parameters.\\
\paragraph{SNet-1 (TARNet, adapted from \cite{shalit2017estimating})}
\begin{equation}
\min_{h_0, h_1, \Phi} \frac{1}{n} \sum^n_{i=1} L(h_{W_i}(\Phi(X_i)), Y_i) + \lambda \mathcal{R}(h_0, h_1)
\end{equation}
where $\Phi$ is the shared representation, $h_0$ and $h_1$ are the potential outcome hypothesis functions, $L(\cdot)$ refers to the squared loss if $Y$ is continuous and cross-entropy if $Y$ is binary, and $\mathcal{R}(\cdot)$ is an L2-regularisation term. 

\begin{remark}[Balanced Representations and Causal Identifiability]
Inspired by ideas from domain adaptation, \cite{shalit2017estimating} show that learning invertible feature maps $\Phi$ that minimize the distance between treatment groups in feature space leads to minimization of a generalization error bound. Nonetheless, we refrain from using \textit{balanced} representations here (and hence rely on \cite{shalit2017estimating}'s TARNet instead of their proposed counterfactual regression method based on balanced representations (CFR)), because minimizing \cite{shalit2017estimating}'s proposed loss function associated with CFR does not necessarily result in invertible representations (\cite{johansson2019support}, \cite{zhang2020learning}) -- which can be detrimental in the causal inference setting. If information is discarded by artificially balancing treatment groups in a new feature space, this can re-introduce selection bias (which was controlled for in the original feature space!). While we do not investigate this line of thought further here, it would be possible to add an invertibility-penalty similar to the one used in \cite{zhang2020learning} to \cite{shalit2017estimating}'s CFR loss function to circumvent this problem. We also note that representations do not necessarily have to be invertible for causal identification -- rather, representations have to preserve all identifying conditional independence relationships. Formally, $\Phi(X)$ has to satisfy $W \indep X | \Phi(X)$ for confounders $X$ which means that it should take the role of a balancing score \citep{rosenbaum1983central}. Hence, it would not be problematic to discard features \textit{without} confounding effect, even though this leads to non-invertible representations. 
\end{remark}

\paragraph{SNet-2 (DragonNet, adapted from \cite{shi2019adapting})}
\begin{equation}
\min_{h_0, h_1, \Phi} \frac{1}{n} \sum^n_{i=1} [L(h_{W_i}(\Phi(X_i)), Y_i) + CrossEntropy(h_\pi(\Phi(X_i)), W_i)]+ \lambda \mathcal{R}(h_0, h_1, h_\pi)
\end{equation}
where the only difference with SNet-1 arises from the additional hypothesis function $h_\pi$ for the propensity score, which is also learnt using the representation $\Phi$.

\paragraph{SNet-3 and SNet}
\cite{hassanpour2019learning}'s DR-CFR (SNet-3) is built upon three instead of one representation, where one affects outcome only ($\Phi_O$), one affects treatment only ($\Phi_W$) and one affects both and is hence a true confounder ($\Phi_C$). We use the following adapted loss function to implement SNet-3 and SNet (where SNet uses the same loss function but adds two further representations $\Phi_{\mu_0}$ and $\Phi_{\mu_1}$  which affect only the respective treatment group.)  

\begin{equation}
\begin{split}
\min_{h_0, h_1, \Phi_O, \Phi_C, \Phi_W} \frac{1}{n}\sum^n_{i=1} [L(h_{W_i}(\Phi_O(X_i), \Phi_C(X_i)), Y_i)+ CrossEntropy(h_\pi(\Phi_C(X_i), \Phi_W), W_i)]+ \\ \lambda \mathcal{R}(h_0, h_1, h_\pi) + \gamma \mathcal{R}_O(\Phi_O, \Phi_C, \Phi_W)
\end{split}
\end{equation}
As in \cite{wu2020learning}'s adaptation of DR-CFR, we also add an orthogonalization term $\mathcal{R}_O$ to our implementation, which ensures that each variable in $X$ affects only one of the 3 (5) representations. This is necessary because -- as representations and outcome functions are learned jointly -- the learned representations are not identifiable otherwise.
To enforce separation and hence specialisation of each representation, we add a regularization term that penalizes whenever a variable enters two representations. Let $W^{1, \Phi_k}$ be the first weight matrix in representation $\Phi_k$ such that $X W^{1, \Phi_k}$ is the first pre-activation in the representation layer. Whether variable $j$ enters representation $\Phi_k$ can be measured by $\bar{W}_{\Phi_k, j} = \sum_{u} |W^{1, \Phi_k}_{j,u}|$, and the orthogonalization term  $\mathcal{R}_O$ simply consists of all cross-products $\bar{W}_{\Phi_k, j} \times \bar{W}_{\Phi_l, j}$ of the different representation contributions. While this does not force hard-decomposition, it does penalize a variable entering multiple representations, leading to good disentanglement in practice.

\subsection{IMPLEMENTATION DETAILS}
In our implementations, we use components similar to those used in \cite{shalit2017estimating} for all networks. In particular, we use dense layers with exponential linear units (ELU) as nonlinear activation functions. We train with Adam \citep{kingma2014adam}, minibatches of size 100, and use early stopping based on a 30\% validation split. For SNet-1 and SNet-2, the representation $\Phi$ consists of 3 layers with 200 units, while for SNet-3 $\Phi_C$ has 150 units and $\Phi_O$ and $\Phi_W$ have 50. In SNet, $\Phi_C$ and $\Phi_W$ have 100 units, while $\Phi_O$, $\Phi_{\mu_0}$ and $\Phi_{\mu_1}$ have 50 units. For hypothesis functions without shared layers (TNet, the second step regressions, and the propensity score in SNet-1), each hypothesis function gets 3 layers of 200 units of its own. Finally, each hypothesis function (output head) consists of 2 additional layers with 100 units and a final prediction layer (with sigmoid-activation for the propensity score). This set-up ensures that each estimated function ($\hat{\mu}_w(x)$, $\hat{\pi}(x)$ and $\hat{\tau}(x)$) has access to the same amount of layers and units in total, and each architecture can hence represent equally complex nuisance functions. We set $\lambda=0.0001$ throughout, and $\gamma=0$ in the IHDP experiments and $\gamma=0.01$ in the simulation study. All models were implemented using jax \citep{jax2018github}. Sklearn-style implementations for all models are available at  \url{https://github.com/AliciaCurth/CATENets} and at \url{https://github.com/vanderschaarlab/mlforhealthlabpub/tree/main/alg/CATENets}. 

\section{EXPERIMENTS}
The code used to perform all experiments is available at \url{https://github.com/AliciaCurth/CATENets} and at \url{https://github.com/vanderschaarlab/mlforhealthlabpub/tree/main/alg/CATENets}.
\subsection{Simulation set-up}
We use a simulation set-up that is partially inspired by that used in \cite{hassanpour2019learning}.  In particular, across all settings, we use $d=25$ multivariate normal covariates $X$ which are simulated in disjoint subsets $X_s$ of $X$, with size $d_s$ according to $X_s \sim \mathcal{N}(0, I)$. For each setting and each training sample size $n \in \{1000, 2000, 5000, 10000\}$ we draw 10 independent training samples of size $n$ and test samples of size $500$.

For settings (i) and (ii), we use covariates $X_C$, which are confounders affecting both outcome and treatment assignment, and $X_O$, affecting only outcomes. Both $X_C$ and $X_O$ are composed of $5$ covariates. We model the baseline outcome as
\begin{equation}
\mu_0(x) = \mathds{1}^\top X_{CO}^2
\end{equation}
where $ X_{CO}= [X_C, X_O]$, $\mathds{1}$ is the unit vector and $X_{CO}$ is squared elementwise. 

Treatments are sampled as a Bernoulli random variable using the propensity score:
\begin{equation}
\pi(x) = expit(\xi (\frac{1}{d_c} \mathds{1}^\top X^2_c - \omega))
\end{equation}
where $\xi$ determines the extent of the selection bias. In our experiments we set $\xi=3$. Further, we adaptively set $\omega=median(\frac{1}{d_c} \mathds{1}^\top X^2_c)$ in each simulation run to center propensity scores (if we would not do so, the squares in the specification would lead to a much larger treatment group than control group) . 

In setting (i), where there is no treatment effect, we set $\mu_{1, (i)}(x) = \mu_0(x)$. In setting (ii), we use 5 additional covariates $X_\tau$ to model a treatment effect:
\begin{equation}
\mu_{1, (ii)}(x) = \mu_0(x) + \mathds{1}^\top X_\tau^2
\end{equation}

For setting (iii), we simulate a setting without confounding ($\pi(x)=0.5$) where the potential outcome functions are determined by non-overlapping covariate sets, both of dimension 10, i.e.
\begin{equation}
\mu_{0, (iii)} = \mathds{1}^\top X_{\mu_0}^2 \text{ and } \mu_{1, (iii)} =  \mathds{1}^\top X_{\mu_1}^2
\end{equation}

Finally, in all three settings we compute outcomes as
\begin{equation}
Y_i = W_i \mu_1(X_i) + (1-W_i)\mu_0(X_i) + \epsilon_i
\end{equation}
with $\epsilon_i \sim \mathcal{N}(0,1)$

\subsection{IHDP data-set}
We use an adapted version of the Infant Health and Development Program (IHDP) benchmark used in \cite{shalit2017estimating} and extensions, created by \cite{Hill2011}. The underlying data-set belongs to a real randomized experiment targeting premature infants with low birth weight with an intervention, which contains 25 covariates (6 continuous and 19 binary) capturing aspects related to children and their mothers. The benchmark data-set was created by excluding a non-random proportion of treated individuals, namely those with nonwhite mothers. The final data-set consists of 747 observations (139 treated, 608 control), and overlap is not satisfied (as $\pi(x)$ is not necessarily non-zero for all observations in the control group). While the covariate data is real, the outcomes are simulated according to setting ``B" described in \cite{Hill2011}, which satisfies $Y(0) \sim \mathcal{N}(exp((X+W)\beta), 1)$ and $Y(1) \sim \mathcal{N}(X\beta - \omega, 1)$ with $W$ an offset matrix and the coefficient $\beta$ has entries in $(0, 0.1, 0.2, 0.3, 0.4)$, where each entry is independently sampled with probabilities (0.6, 0.1, 0.1, 0.1, 0.1). We use the 100 repetitions of the simulation provided by \cite{shalit2017estimating}.

\paragraph{Rescaled data-set} Rooted in the simulation specification used to obtain the IHDP regression surfaces, we observed that the scale of CATE varied by orders of magnitude across different runs of the simulation, making the RMSE incomparable across runs. We found that by averaging across the data-sets, the relative performance was dominated by runs with high variance in CATE (which are those where many variables have the larger non-zero coefficients), distorting the per-run relative performance we observed. Therefore, we decided to rescale the outcomes of runs where the training set had $\sigma^2_{CATE}>1$. For these runs, we kept the original error terms $\epsilon_i = Y_i(w) - \mu_{w,i}$ (which were $\mathcal{N}(0,1)$ across all runs) but rescaled the expected potential outcomes as $\tilde{\mu}_{w, i} = \frac{\mu_{w,i}}{\sigma_{CATE}}$ before adding back the error-term. 
\begin{figure}\label{catevar}
\includegraphics[width=\textwidth]{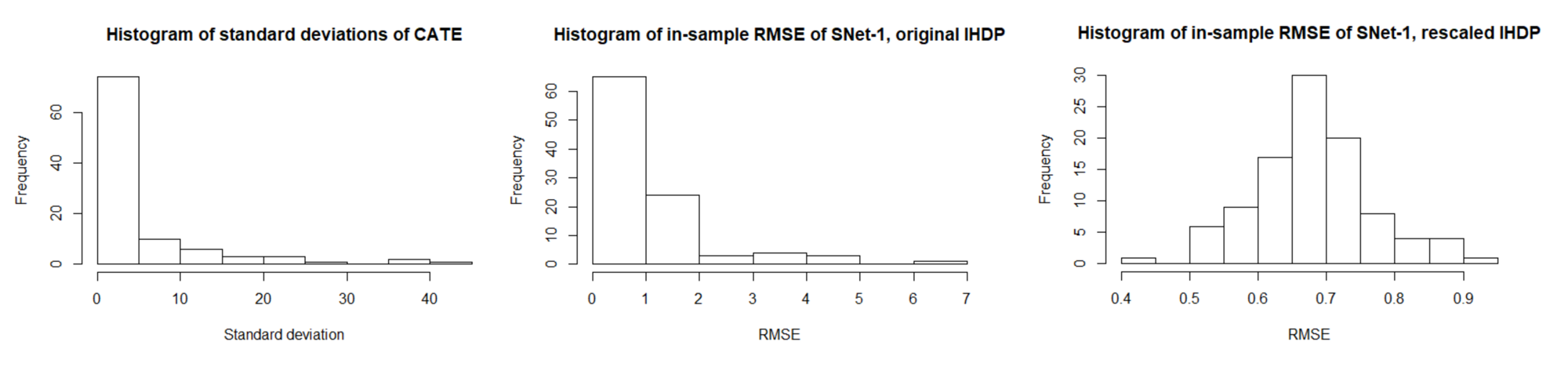}
\caption{From left to right: Histograms of the standard deviation of CATE, in-sample RMSE of SNet-1 on the original IHDP data-set and the rescaled data-set}
\end{figure}
In Figure \ref{catevar}, we plot $\sigma_{CATE}$ as well as the distribution of RMSE in the original and the adapted version of the data-sets for SNet-1, illustrating that only after rescaling the data-set the RMSE results in comparable (and even approximately normal) performance across runs.

\end{document}